\documentclass[preprint,12pt]{elsarticle}

\usepackage{amssymb}
\usepackage{amsthm}
\usepackage{amsmath}
\usepackage[
top    = 1in,
bottom = 1in,
left   = 1in,
right  = 1in]{geometry}

\usepackage{lineno}
\usepackage{xcolor}
\usepackage{lipsum} 
\usepackage{color,soul}
\usepackage[normalem]{ulem}
\usepackage{caption}
\usepackage{tabularx}
\usepackage{graphicx}
\usepackage{array}
\usepackage{enumitem}
\usepackage{multirow}
\usepackage{makecell}
\usepackage{longtable}

\journal{Expert Systems with Applications}

\begin{document}

\begin{frontmatter}



\title{An Interpretable Deep Hierarchical Semantic Convolutional Neural Network for Lung Nodule Malignancy Classification}


\author[1,2]{Shiwen Shen}
\ead{shiwenshen@engineering.ucla.edu}
\author[1,2]{Simon X. Han}
\ead{simon.x.han@ucla.edu}
\author[1,2]{Denise R. Aberle}
\ead{daberle@mednet.ucla.edu }
\author[2]{Alex A.T. Bui}
\ead{buia@mii.ucla.edu}
\author[2]{Willliam Hsu\corref{co1}}
\ead{whsu@mednet.ucla.edu}

\address[1]{Department of Bioengineering, University of California, Los Angeles, CA, USA}
\address[2]{Medical Imaging \& Informatics Group, Department of Radiological Sciences, University of California, Los Angeles, CA, USA}
\cortext[co1]{Medical Imaging \& Informatics, 924 Westwood Boulevard, Suite 420, Los Angeles, CA 90024, USA.}

\begin{abstract}
While deep learning methods are increasingly being applied to tasks such as computer-aided diagnosis, these models are difficult to interpret, do not incorporate prior domain knowledge, and are often considered as a ``black-box.'' The lack of model interpretability hinders them from being fully understood by target users such as radiologists. In this paper, we present a novel interpretable deep hierarchical semantic convolutional neural network (HSCNN) to predict whether a given pulmonary nodule observed on a computed tomography (CT) scan is malignant. Our network provides two levels of output: 1) low-level radiologist semantic features; and 2) a high-level malignancy prediction score. The low-level semantic outputs quantify the diagnostic features used by radiologists and serve to explain how the model interprets the images in an expert-driven manner. The information from these low-level tasks, along with the representations learned by the convolutional layers, are then combined and used to infer the high-level task of predicting nodule malignancy. This unified architecture is trained by optimizing a global loss function including both low- and high-level tasks, thereby learning all the parameters within a joint framework. Our experimental results using the Lung Image Database Consortium (LIDC) show that the proposed method not only produces interpretable lung cancer predictions but also achieves significantly better results compared to common 3D CNN approaches. 
\\
\end{abstract}


\end{frontmatter}

\section{Introduction and Background}
\label{Intro}

\begin{figure}[ht!]
	\centering
	\includegraphics[scale=0.75]{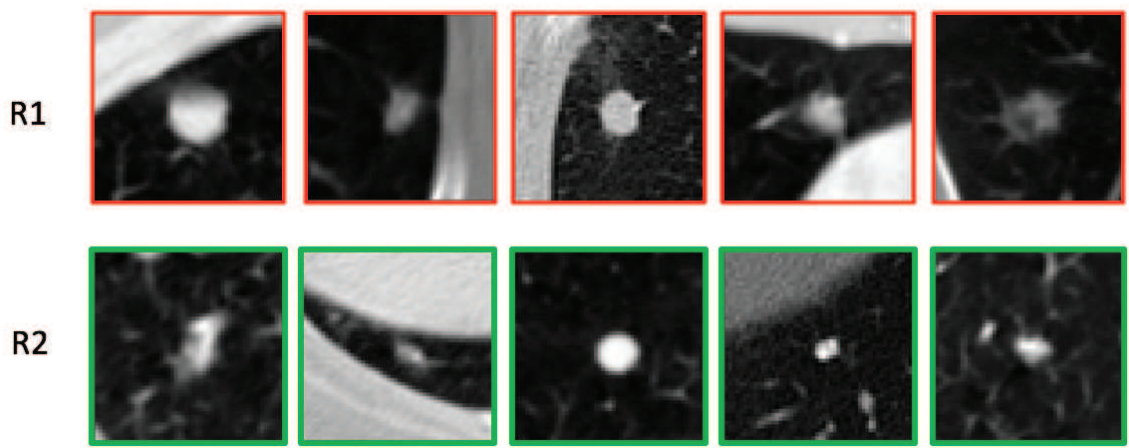} 
	\caption{Illustrations of malignant and benign nodules: R1 are malignant nodules; R2 are benign nodules.}
	\label{f_malig_example}
\end{figure}

Lung cancer is the leading cause of cancer mortality worldwide \cite{torre2016lung, shen2017bayesian}. Computed tomography (CT) imaging is widely used to detect pulmonary nodules and forms the basis for diagnosing lung cancer. The landmark National Lung Screening Trial (NLST) \cite{national2011reduced} in the United States demonstrated a 20\% lung cancer mortality reduction in high-risk subjects who underwent screening using low-dose CT (relative to using plain chest radiography). Based on the findings of the NLST, the United States Preventative Services Task Force (USPSTF) went on to recommend low-dose CT lung cancer screening for current and former smokers aged 55-80 with a smoking history of at least 30 pack-years, or former smokers having quit within the past 15 years \cite{ten2017risk}. However, the potential consequences of implementing lung cancer screening is an increase in false positive screens that result in unnecessary medical, economic, and psychological costs. Indeed, some studies indicate that the false positive rate for low-dose CT is upwards of 20\%; less experienced radiologists have highly variable detection rates, particularly in subtle cases, as interpretation heavily relies on past experience \cite{zhao2013exploring}. Figure \ref{f_malig_example} shows examples of malignant and benign nodules, helping to illustrate the challenges of differentiating between the two groups. 

In response, computer-aided diagnosis (CADx) systems \cite{armato2003automated, shen2015automated, duggan2015technique, firmino2016computer, amir2016after, huang2017added} have been explored to assist radiologists in the interpretation process; these help to separate malignant from benign nodules and show promise in increasing the positive predictive value and reducing the false positive rates in characterizing small nodules \cite{huang2017added}. Broadly, a contemporary lung nodule CADx system comprises three components: 1) nodule segmentation; 2) feature extraction from nodule candidates; and 3) classification of the candidate as benign or malignant based on its extracted features. Segmentation of the lung nodule is the first step to identify pertinent regions of interest (ROIs) so as to focus the ensuing analyses. In the second step, computed features such as texture, morphology, and gray-level statistics are extracted to characterize the nodule across various visual and spatial dimensions. Finally, using the computed features, the nodule candidates are categorized into benign or malignant by a classifier such as support vector machine, random forest, or gradient boosting tree. For example, Armato et al. \cite{armato2003automated} segmented the lung nodule using multilevel thresholding techniques; extracted morphological and gray-level features; and classified nodules as benign or malignant using linear discriminant analysis. Zinovev et al. \cite{zinovev2011probabilistic} employed both texture and intensity features using belief decision trees and a multi-label approach to perform lung nodule classification. Way et al. \cite{way2009computer} segmented lung nodules using k-means clustering, combined nodule surface features together with texture and morphological features, and used linear discriminant analysis to diagnose malignant lung cancers. Notably, the prevalent portion of these extracted features, such as volume and shape, are sensitive to the underlying variability arising from the lung nodule segmentation results. This variability may be caused by: 1) the challenges inherent in nodule delineation due to the range of nodule morphology and connectivity complexity \cite{shen2015multi}; and 2) the variation posed by the computational segmentation models themselves in capturing quantitative characteristics \cite{shen2017multi, piedra2016assessing}. Thus, using segmented regions may lead to inaccurate features that are subsequently used as inputs into downstream classifiers \cite{shen2017multi}. Another critical question raised by this type of CADx design is how to define the ``optimal'' subset of features that can best encode characteristics of the lung nodule \cite{ciompi2015automatic}. The optimal feature set is dependent on the characteristics of the training dataset, and feature selection and classification methods, which makes achieving comparable results using different datasets difficult.

To overcome these issues, deep learning methods \cite{shen2015multi, shen2017multi, ciompi2015automatic, kumar2015lung, hua2015computer}, particularly convolutional neural networks (CNNs), have recently been used for lung nodule classification, with some promising results. These deep learning models prove capable of adaptively learning image representations in a fully data-driven way, taking raw image data as input without relying on \textit{a priori} nodule segmentation masks or handcrafted feature designs. For instance, Kumar et al. \cite{kumar2015lung} first trained an unsupervised deep autoencoder to extract latent features from 2D CT patches. These extracted deep features were then used together with decision trees to predict lung cancer. Hua et al. \cite{hua2015computer} employed supervised techniques with a deep belief network and CNN to train models to classify lung nodules as benign or malignant. Their models outperformed two baseline methods: using scale-invariant feature transform (SIFT) features and local binary patterns (LBP) \cite{farag2011evaluation}; and using fractal analysis \cite{lin2013automatic}. Ciompi et al. \cite{ciompi2015automatic} used pre-trained CNN models to classify candidates as peri-fissural nodules (PFNs) or non-PFNs. Deep features were extracted from the pre-trained model for three 2D image patches in axial, coronal, and sagittal views. An ensemble of the deep features and a bag of frequency features were then used to train supervised binary classifiers for the PFN classification task. Shen et. al. \cite{shen2015multi} designed a multi-scale CNN using 3D nodule patches at three scales to perform the lung cancer diagnosis task. This work is further extended in \cite{shen2017multi} by adding a multi-crop pooling strategy to improve model performance. Markedly, these cited works use deep learning as a ``black-box'' and are not able to explain what representations have been learned or why the model generates a given prediction. This low degree of interpretability arguably hinders domain experts, such as radiologists, from understanding how the models work and ultimately impedes model adoption for clinical usage. As discussed in \cite{jorritsma2015improving}, improved interpretability is helpful to improve the radiologist-CAD interaction to allow radiologists to calibrate their trust in the CAD system. Moreover, human domain knowledge performs well across a wide range of observed nodule morphologies to differentiate benign from malignant nodules. Nonetheless, these domain knowledge are presently not incorporated into deep learning frameworks. 

A number of radiologist-interpreted features derived from CT scans have been considered influential when assessing the malignancy of a lung nodule \cite{kim2015quantitative, erasmus2000solitary}. These features are referred to as diagnostic \textit{semantic} features in this study. Examples of such semantic features include nodule spiculation, lobulation, consistency (texture) and shape. Although qualitative in nature, studies have shown that these semantic features can be characterized numerically using low-level image features \cite{kaya2015weighted}. Hancock et al. \cite{hancock2016lung}  demonstrated that machine learning can achieve high prediction accuracy for lung cancer malignancy using only semantic features as inputs. In addition, semantic features are intuitive to radiologists and are moderately robust against perturbations in image resolution and reconstruction kernel. An opportunity exists to incorporate these semantic features into the design of deep learning models, combining the advantages of both.

In this study, we propose a novel interpretable hierarchical semantic convolutional neural network (HSCNN) to predict whether a nodule is malignant in CT images. The HSCNN takes the raw CT image cubes centered at nodules as input and generates two levels of outputs. The first predictive level provides intermediate outputs in terms of diagnostic semantic features, while the second level represents the final lung nodule malignancy prediction score. Jump connections are employed to feed the information learned from the first level semantic features to the final malignancy prediction. As such, our first level outputs provide explanations about what the HSCNN model has learned from the raw image data and correlates semantic features with the specific malignancy prediction; it also provides additional information to improve the final malignancy prediction task through the jump connections. This unified model is trained as a whole by minimizing a global cost function, where both first- and second-level task losses are included.

The contributions of this paper are fourfold:

\begin{enumerate}
  \item We describe an approach to build a radiologist-interpretable deep convolution neural network. The intermediate outputs from the model give predictions of diagnostic semantic features associated with the final classification, helping to explain the prediction. To the best of our knowledge, this is the first example of a network architecture emphasizing the interpretability of the results.
  \item We provide a hierarchical design that integrates both semantic features and deep features to predict malignancy. Shared convolution modules in the HSCNN are used to learn generalizable features across tasks. The information learned for each specific low-level semantic feature is then fed into the final high-level malignancy prediction task.
  \item We present a new global cost function to train the whole model jointly, taking both first- and second-level outputs into consideration simultaneously. The new objective function concurrently handles data imbalance issues for both tasks.
  \item We show how semantic features can be quantified using a deep CNN model from reconstructed CT images in a data-driven fashion. The produced model generates labels for all semantic features on previously unseen cases.
\end{enumerate}

The remainder of this paper is organized as follows. In Section
\ref{sec:Materials}, we describe the dataset used in this study and
the proposed HSCNN model. In Section \ref{hscnn_results}, we
present results and compare the proposed method with a traditional 3D
CNN. In Sections \ref{discuss_results} \& \ref{conclustion_hscnn}, we discuss the findings and
limitations of the work. 

\section{Materials and Methods}
\label{sec:Materials}
\subsection{Lung Image Database Consortium Dataset}
\label{Materials:2.1}

The Lung Image Database Consortium image collection (LIDC-IDRI) \cite{armato2011lung} is a publicly available dataset, which we used to train and test our proposed methods. LIDC-IDRI contains both screening and diagnostic CT scans collected from 7 academic centers and 8 medical imaging companies. Inclusion criteria for CT scans were: 1) having a collimation and reconstruction interval no greater than 3 mm; and 2) each scan approximately containing no more than 6 lung nodules with the longest dimension ranging from 3-30 mm, as determined by a cursory review during case selection at the originating institution \cite{armato2011lung}. For the whole dataset, the slice thicknesses varied from 0.6 to 5 mm, and the in-plane pixel size varied from 0.461 to 0.977 mm. LIDC-IDRI comprises 1,018 cases (representing 1,010 different patients, 8 patients having 2 distinct scans), with each including images from a clinical CT scan and an associated eXtensible Markup Language (XML) file. The XML files record the reference standard for lung nodule locations, as manually annotated by four radiologists following a two-phase image annotation process. Pixel-level 3D contour segmentations, panel opinions on the assessment of nodule likelihood for malignancy, and eight nodule characteristics were generated only on lesions categorized as nodules $\geq$ 3 mm. The eight nodule characteristics are semantic diagnostic features, including: calcification, subtlety, lobulation, sphericity, internal structure, margin, texture, spiculation, and malignancy. Each feature is rated from 1 to 5 or 6 by radiologists. Table \ref{table:semantic labels} lists the description and definitions for each of the labels from \cite{mcnitt2007lung}.

\begin{longtable}[ht!]{ >{\small} p{3.5cm} >{\small} p{6 cm}  >{\small} p{5 cm} }

\caption{Detailed nodule characteristics labels in LIDC dataset.} \\
\hline
\textbf{Semantic Feature} & \textbf{Description} & \textbf{Ratings} \\
\hline
\endfirsthead

\hline
\textbf{Semantic Feature} & \textbf{Description} & \textbf{Ratings} \\
\hline
\endhead

\hline
\endfoot

\hline
\endlastfoot

Malignancy & Likelihood of malignancy &  \makecell[l]{1. Highly unlikely\\2. Moderately unlikely \\3. Indeterminate \\4. Moderately suspicious \\5. Highly suspicious} \\
\hline
 Margin & How well defined the margins are & \makecell[l]{1. Poorly defined \\2. \\3.  \\ 4.  \\ 5. Sharp}\\
\hline
Sphericity & Three dimensional shape in terms of roundness & \makecell[l]{1. Linear \\2. \\3. Ovoid \\ 4.  \\ 5. Round}\\
\hline
Subtlety & Difficulty of detection relative to surround & \makecell[l]{1. Extremely subtle \\2. Moderately subtle \\3. Fairly subtle \\ 4. Moderately obvious \\ 5. Obvious}\\
\hline
Spiculation & Degree of exhibition of spicules & \makecell[l]{1. Marked \\2. \\3.  \\ 4.  \\ 5. None}\\

\hline
Radiographic solidity (texture) & Internal texture (consistency) of nodule & \makecell[l]{1. Non-solid \\2. \\3. Part Solid \\ 4.  \\ 5. Solid}\\
\hline
Calcification & Presence and pattern of calcification & \makecell[l]{1. Popcorn \\2. Laminated \\3. Solid \\4. Non-central \\5. Central \\6. Absent}\\
\hline
Internal structure & Expected internal composition of the nodule & \makecell[l]{1. Soft tissue \\2. Fluid \\3. Fat \\ 4.  \\ 5. Air}\\
\hline
Lobulation & The presence and degree of lobulation of the nodule margin & \makecell[l]{1. Marked \\2. \\3.  \\ 4.  \\ 5. None}\\
\label{table:semantic labels}
\end{longtable}

\subsection{Our Usage of the LIDC dataset}

\label{Materials_lidc_use}

\begin{table}[h]
\centering
\caption{Summary of generating binary labels from LIDC rating scales for nodule characteristics.}
\label{hscnn_label_schema}
\begin{tabular}{c|c|c}
\hline
\textbf{Nodule characteristics} & \textbf{Label 0}   & \textbf{Label 1} \\ \hline
Malignancy                      & \begin{tabular}[c]{@{}c@{}}Scale 1 - 3\\ Benign\end{tabular}                                           & \begin{tabular}[c]{@{}c@{}}Scale 4 - 5\\ Malignant\end{tabular}     \\ \hline
Sphericity                      & \begin{tabular}[c]{@{}c@{}}Scale 1 - 3\\ Lesser roundness\end{tabular}                                 & \begin{tabular}[c]{@{}c@{}}Scale 4 - 5\\ High degree of roundness\end{tabular}  \\ \hline
Margin                          & \begin{tabular}[c]{@{}c@{}}Scale 1 - 3\\ Poorly defined margin\end{tabular}                            & \begin{tabular}[c]{@{}c@{}}Scale 4 - 5\\ Sharp margin\end{tabular}  \\ \hline
Subtlety                        & \begin{tabular}[c]{@{}c@{}}Scale 1 - 3\\ Poor contrast between nodule \\ and surroundings\end{tabular} & \begin{tabular}[c]{@{}c@{}}Scale 4 - 5\\ High contrast between nodule \\ and surroundings\end{tabular} \\ \hline
Texture                         & \begin{tabular}[c]{@{}c@{}}Scale 1 - 3\\ Non-solid internal density\end{tabular}                       & \begin{tabular}[c]{@{}c@{}}Scale 4 - 5\\ Solid internal density\end{tabular} \\ \hline
Calcification                   & \begin{tabular}[c]{@{}c@{}}Scale 1 - 5\\ Present of calcification\end{tabular}                         & \begin{tabular}[c]{@{}c@{}}Scale 6\\ Absent of calcification\end{tabular} \\ \hline                             
\end{tabular}
\end{table}

\begin{figure}[h]
	\centering
	\includegraphics[scale=0.4]{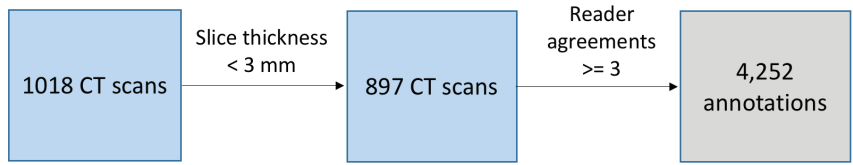} 
	\caption{Lung nodule annotation selecting criteria.}
	\label{f_selecting_rules}
\end{figure}

In the LIDC, nodules could be associated with 1-4 annotations, depending on how many of the four readers completed the annotation task. To determine which annotations referred to the same nodule, we used an annotation list provided in \cite{lidc_size_2011}. Only nodules identified by at least three radiologists on CT scans with slice thicknesses smaller than 3 mm were included in this study, as shown in Figure \ref{f_selecting_rules}. The inclusion criteria resulted in sub-selecting 4,252 nodule annotations. Each annotation is considered independently (e.g., one object might be marked by all four radiologists as a nodule and was considered as four independent nodules), the rationale for this assumption being that the exact boundaries of a given nodule lacks universal agreement. Furthermore, following the conventions used by others \cite{clark2013cancer, hancock2016lung, froz2017lung}, we wished to maximize use of existing annotations. Uniform labels are assigned to all annotations referring to the same nodule for each feature. As shown in Table \ref{table:semantic labels}, the LIDC annotation process employed ordinal scales to label likelihood of malignancy and four semantic features: margin, sphericity, nodule subtlety, and texture (consistency). Scores for these five nodule characteristics were binarized by averaging the scores for each nodule as in \cite{shen2015multi} and assigning a threshold at 4 to distinguish scores of 1-3 (Label 0) from scores of 4-5 (Label 1). Label 0 typically indicates a benign nodule, poorly defined margin, lesser roundness, poor conspicuity between nodule and surroundings, and a non-solid (ground-glass-like) consistency. Conversely, Label 1 more typically denotes a malignant nodule, sharp margins, higher sphericity, high conspicuity between nodule and surroundings, and solid consistency. Calcification was annotated using a categorical scale from 1 to 6 in which 6 refers to the absence of calcification; here, we averaged ratings for each nodule by majority vote per \cite{clark2013cancer}. Nodules with average ratings of 6 were labeled as absent of calcification pattern (Label 1); all other ratings represented the presence of calcification (Label 0). 

\begin{table}[h!]
\caption{Label counts for nodule characteristics.}
	\centering
	\begin{tabular}{c |c c c}
		\hline
		Nodule characteristics		& Label 0 (\#)	& Label 1 (\#)	& Total (\#) \\
		\hline
		Malignancy				& 3212 	& 1040 	& 4252 \\
		Sphericity					& 2304	& 1948	& 4252\\
		Margin					& 1640	& 2612	& 4252\\
		Subtlety					& 1570	& 2682	& 4252\\
		Texture					& 518	& 3734	& 4252\\
		Calcification				& 496	& 3756	& 4252\\
		\hline
	\end{tabular}
	\label{table:label_counts}
\end{table}

The feature "internal structure" was overwhelmingly annotated as soft tissue, thus provided little discriminative information \cite{hancock2016lung} and was excluded from our analysis. Moreover, the Cancer Imaging Archive (TCIA) reported that an indeterminate subset of cases in the dataset were inconsistently annotated with respect to spiculation and lobulation \cite{lidc_annotation_error}. As such, we did not consider these two features in our model.  Finally, it should be noted that the actual diagnosis of the nodules is not known in the LIDC dataset. For the purposes of this work, the likelihood of malignancy served as the proxy for truth. Table \ref{hscnn_label_schema} summarizes the generation of the binary labels from LIDC rating scales as described above. Table \ref{table:label_counts} lists the data counts for each label of the nodule characteristics.

\subsection{Data Preprocessing}
\label{sec:prepro}

The LIDC dataset contains a heterogeneous set of scans obtained using various acquisition and reconstruction parameters. To normalize pixel values, all CT scans were first transformed to Hounsfield (HU) scales using the information in the DICOM (Digital Imaging and Communication in Medicine) series header and converted to a range of $(0, 1)$ from (-1000, 500 HU). A 3D cube of 40 $\times$ 40 $\times$ 40 mm were extracted for each candidate. Each cube was centered around the candidate. 40 mm was chosen so that all candidates would be fully contained in this cube as the largest nodules in our subset were maximally 30 mm. We then rescaled each cube to a fixed size of pixels in all three dimensions, resulting in isotropic cubes for all cases. This preprocessing method retained the original relative nodule size information, which is considered useful domain knowledge for the the following tasks.

\subsection{Hierarchical Semantic Convolutional Neural Network}
\label{sec:hscnn}

\begin{figure}[ht!]
	\centering
	\includegraphics[scale=0.42]{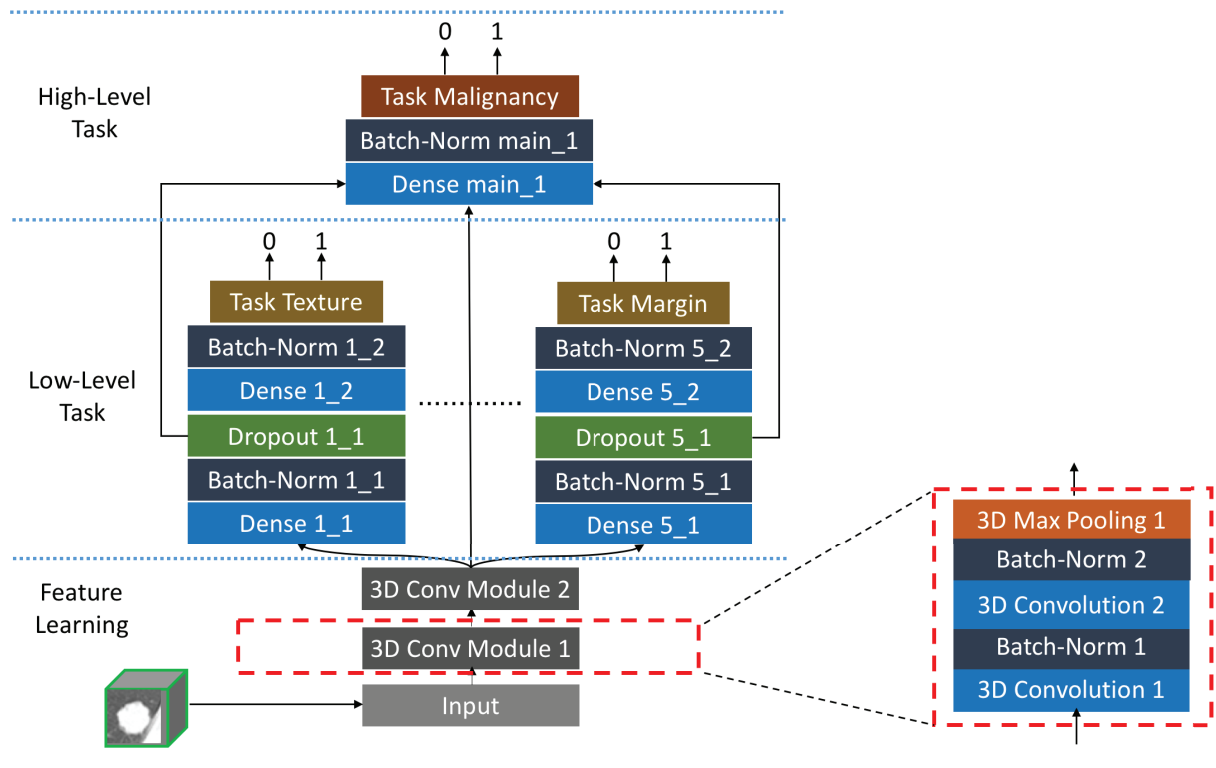} 
	\caption{Model architecture of the hierarchical semantic convolutional neural network.}
	\label{f_hscnn_frame}
\end{figure}

The proposed HSCNN utilizes a 3D patch capturing the lung nodule as input and outputs two levels of predictions, as shown in Figure \ref{f_hscnn_frame}. This architecture comprises three parts: 1) a feature learning module; 2) a low-level task module; and 3) a high-level task module. The feature learning component adaptively learns the image features that are generalizable across different tasks. The low-level task predicts five semantic diagnostic features: margin, texture, sphericity, subtlety, and calcification. The high-level task incorporates information from both the generalizable image features and the low-level tasks to produce an overall prediction of lung nodule malignancy. 

The feature learning component (Figure \ref{f_hscnn_frame}, feature learning) is the first processing unit of the proposed network. It consists of two convolution module blocks where each block shares the same structure and contains two stacked 3D convolution layers followed by batch normalization and one 3D average pooling layer. Each convolution layer has a kernel size of $3 \times 3 \times 3$. These layers perform the convolution operation on input feature maps along all three dimensions of the input cube to produce an output feature map defined by:

\begin{equation}
f^{j}=\sum_{i} c^{j} \ast f^{i} + b^{j}
\end{equation}

\noindent where $f^{j}$ and $f^{i}$ are the $j^{th}$ output feature map and $i^{th}$ input feature map, respectively. And $c^{j}$ is the $j^{th}$ convolution kernel and $\ast$ represents the 3D convolution operation between the convolution kernel and input feature map. $b^{j}$ is the $j^{th}$ bias corresponding to the $j^{th}$ convolution kernel. After convolution, batch normalization is applied to all output feature maps to accelerate the training process and reduce the internal covariate shift by normalizing the feature maps \cite{ioffe2015batch}. Rectified linear units (ReLUs) \cite{krizhevsky2012imagenet} are used as the nonlinear activation functions to take the output from batch normalization. 16 feature maps are used for both convolution layers in the first convolution module, and 32 feature maps are adopted for both convolution layers in the second convolution module. A 3D max pooling layer is used in the end for each convolution module block to progressively reduce the spatial size of the feature maps to reduce the number of parameters and control for overfitting. This layer is defined as:

\begin{equation}
\begin{split}
\hat{f}^{i}_{x,y,z}=max\{f^{i}_{x',y',z'};x'\in[x\cdot s_{x}, \; x\cdot s_{x} + d_{x} -1], \\
y'\in[y\cdot s_{y}, \; y\cdot s_{y} + d_{y} -1], \\
 z'\in[z\cdot s_{z}, \; z\cdot s_{z} + d_{z} -1] \}
 \end{split}
\end{equation}

\noindent where $x$ (the row index), $y$ (the column index), and $z$ (the depth index) start from zero. Here, $s$ is the stride size (downscale factor) and $d$ is the size of the max pooling window. We employ a pooling window size of $d = (2, 2, 2)$ and stride size of $s=(2, 2, 2)$. This design downsamples the input feature maps by a factor of 2 across all three cube dimensions. This pooling layer has no learnable parameters. 

After the last convolutional module, output features are fed simultaneously into the low- and high-level task components. The low-level task components (Figure \ref{f_hscnn_frame}, low-level task) consist of five branches, each with the same architecture, addressing a distinct semantic feature task (i.e., texture, margin, sphericity, subtlety, or calcification). A fully-connected layer (densely-connected) is the major basic building block for each of these branches. One fully-connected layer connects each input unit to each output unit, designed to capture correlations from all input feature units to the output. Batch normalization and dropout techniques are both used to control model overfitting. The dropout method randomly removes connections between input and output units during network training to prevent units from co-adapting too much \cite{srivastava2014dropout}. Two fully-connected layers are employed before the final binary prediction with 256 neurons and 64 neurons for the first and second layer, respectively. 

The high-level task component (Figure \ref{f_hscnn_frame}, high-level task) predicts the lung nodule malignancy as the final task. This module concatenates as input the output features from the feature learning component and each of the low-level task branches. As shown in Figure \ref{f_hscnn_frame}, the output feature maps from the last convolution module of the feature learning component are used, along with the output from the last second fully-connected layer of each subtask branch. This design makes the final prediction utilize the basic features learned from the shared convolution modules, and forces the convolution blocks to extract representations that are generalizable across tasks. It also makes use of the information learned from each related explainable subtask to ultimately infer nodule malignancy. The last fully-connected layer in each subtask branch is trained to extract representations more specific to the corresponding subtask compared to the second to last fully-connected layer. Thus, the second to last layer of the subtask branch is chosen to provide less specific but salient information for the final malignancy prediction task. The concatenated features are inputted into a fully-connected layer with 256 neurons, followed by a batch normalization operation before the final malignancy prediction. 

To jointly optimize the the HSCNN during the network training, a global loss function is proposed to maximize the probability of predicting the correct label for each task by:

 \begin{equation}
L_{global}=\frac{1}{N}\sum^{N}_{i=1} (\sum^{5}_{j=1}\lambda_{j} \cdot L_{j, i} + L_{M, i})
\label{hscnn_global_loss}
\end{equation}

\noindent where $N$ is the total number of training samples and $i$ indicates the $i^{th}$ training sample. $j$ is the $j^{th}$ subtask and $j \in [1, 5]$. $\lambda_{j}$ is the weighting hyperparameter for the $j^{th}$ subtask. $L_{j, i}$ represents the loss for sample $i$ and task $j$. $L_{M, i}$ is the loss for the malignancy prediction task for the $i^{th}$ sample. Each loss component is defined as weighted cross entropy loss by:

\begin{equation}
L_{j, i}=-\log{(e^{f_{y_{i}, j}}/\sum_{n}e^{f_{y_{n}, j}})} \cdot \omega_{y_{i}, j}
\end{equation}

\noindent where $y_{i}$ is true label for the $i^{th}$ sample $(x_{i}, y_{i})$. Here, $y_{i}$ equals 0 or 1. $f_{y_{i},j}$ is the prediction score of the true class $y_{i}$ for task $j$ and $f_{y_{n},j}$ represents a prediction score for class $y_{n}$. We use $\omega_{y_{i}, j}$ to represent the weight of class $y_{i}$ for task $j$. The use of $\omega_{y_{i}, j}$ is important because the labels are imbalanced in all the tasks and $\omega_{y_{i}, j}$ is helpful in reducing the training bias introduced by such data imbalance. Specifically, $\omega_{y_{i}, j}$ weights each class loss proportional to the reciprocal of the class counts in the training data. For instance, $\omega_{y_{i}=0, j} = N_{y_{i}=1, j} / (N_{y_{i}=0, j} + N_{y_{i}=1, j})$ and $\omega_{y_{i}=1, j} = N_{y_{i}=0, j} / (N_{y_{i}=0, j} + N_{y_{i}=1, j})$. $N_{y_{i}=1, j}$ represents the total count of samples in the training data for task $j$, where the true class label equals $1$. The global loss function is minimized during the training process by iteratively computing the gradient of $L_{global}$ over the learnable parameters of HSCNN and updates the parameters through back-propagation. During training, model learnable parameters are initialized using the Xavier algorithm \cite{glorot2010understanding} and are updated using the Adam stochastic optimization algorithom \cite{kingma2014adam}.

\subsection{Training}
\label{sec:method_training}

We performed model training, validation, and testing using 897 LIDC cases, selected as described in Section \ref{Materials_lidc_use}. We split these cases into four subsets, where each subset had a similar number of nodules. A 4-fold cross validation study design was employed to obtain the final assessment of the model performance. For each fold, 2 subsets are used for training, 1 subset for validation, and 1 subset for holdout testing. The validation set is used to tune the hyperparameters and test set is employed as external holdout to report the final model performance. Each subset is used as the test set once during the cross validation. This design ensures that the test set is independent of model training and parameter optimizations, and should better reflect the true model performance without information leakage. We note that earlier studies in \cite{shen2015multi, shen2017multi, kumar2015lung, hua2015computer} only use training and validation splits during the cross validation process, without consideration for holdout test sets; such designs arguably have information leakage, and thus tend to over-estimate model performance. 

To better control for model overfitting, 3D data augmentation was applied during the training process. Data augmentation artificially inflates the dataset by using label-preserving transforms to add more invariant data examples and is considered as a model regularization scheme \cite{krizhevsky2012imagenet}. One or more random operations are applied on each training dataset to generate artificial samples. The spatial affine operations used in this study included translating the position of the nodule within 4 mm or flipping the 3D nodule cube along one of the three axes. The translation limit was set to 4 mm to ensure that the boundaries of the largest nodules were captured properly in the 3D cube ($40 \times 40 \times 40$ mm).

\section{Experimental Results}
\label{hscnn_results}

This section first describes how we trained the models. We then compared our model to a traditional 3D CNN model and other state-of-art methods. We then assessed the accuracy of semantic feature predictions, providing illustrations of correct and incorrect predictions.

\subsection{Model Training}
\label{sec:results_implementation}
Models were trained for 300 epochs during each fold of cross-validation. After 100 epochs of training, the model loss on the validation set became stable. The best model for each fold was chosen to be the one that achieved the lowest malignancy prediction loss on the validation dataset. Only the independent test dataset was used to calculate end model performance. An online augmentation scheme was employed during model training: during each training epoch, additional artificially created training samples were generated by randomly picking one or multiple augmentation operations, as described in Section \ref{sec:method_training}. The same augmentation process was also applied to the validation dataset. To capture a majority of nodule morphology while reducing the input data dimensions, the input nodule cube size was set to be $52\times 52 \times 52$ voxels. The learning rate was set to be $0.001$. The convolution kernel size, number of feature maps, pooling window size, downscale factor, and number of neurons for each fully-connected layer were reported in Section \ref{f_hscnn_frame}. The choices of these parameters have been commonly used, as shown in \cite{krizhevsky2012imagenet, Simonyan14c}. The hyperparameters presented in Equation \ref{hscnn_global_loss} were chosen by using a randomized coarse-to-fine grid search with the validation dataset in the first 20 epochs of each fold \cite{bergstra2012random}.

The proposed HSCNN model was implemented in Python 2.7 with TensorFlow \cite{abadi2016tensorflow} and the Keras toolkit \cite{chollet2015keras}. All experiments were performed on a server with 6-core Intel Xeon E5-2630 processor, 32GB memory, and one NVIDIA TITAN Xp GPU (12GB on-board memory). The training of one HSCNN model takes about 5 hours for 300 epochs.

\subsection{Malignancy Prediction Results}
\label{sec:results_results_malig}

\begin{figure}[ht!]
	\centering
	\includegraphics[scale=0.38]{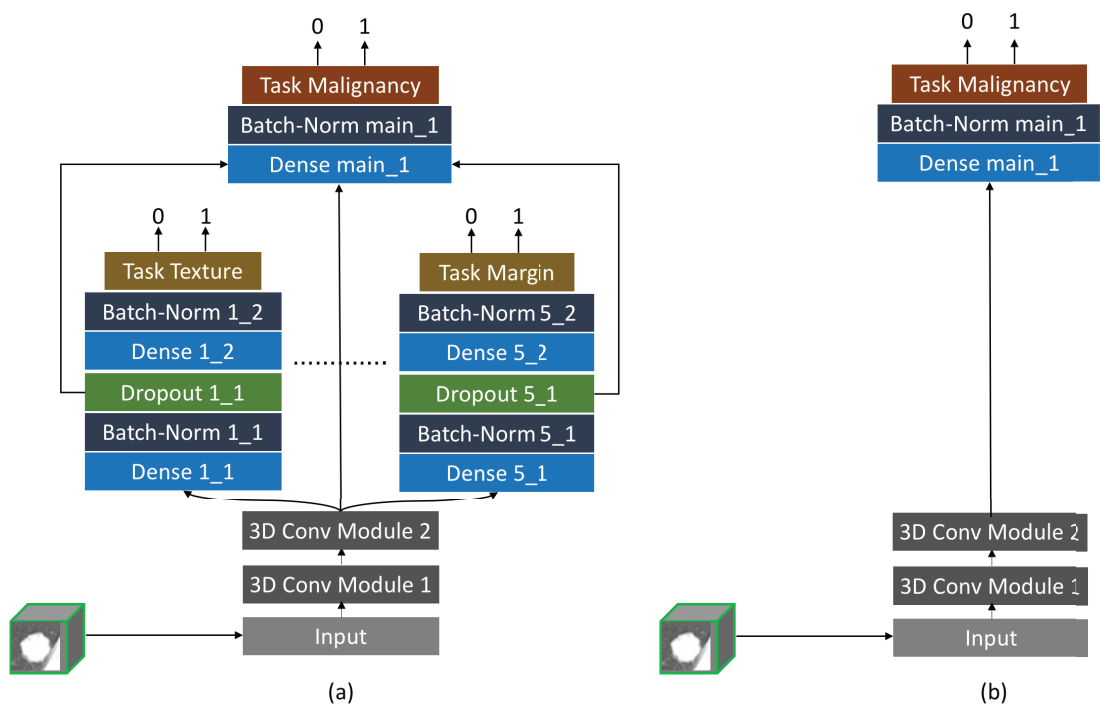} 
	\caption{Framework comparison between proposed HSCNN and baseline 3D CNN. (a) The proposed HSCNN architecture; (b) a baseline 3D CNN architecture. The baseline model has the same structure as the HSCNN but without the low-level semantic task component.}
	\label{f_baseline_hscnn}
\end{figure}

\begin{figure}[t!]
	\centering
	\includegraphics[scale=0.52]{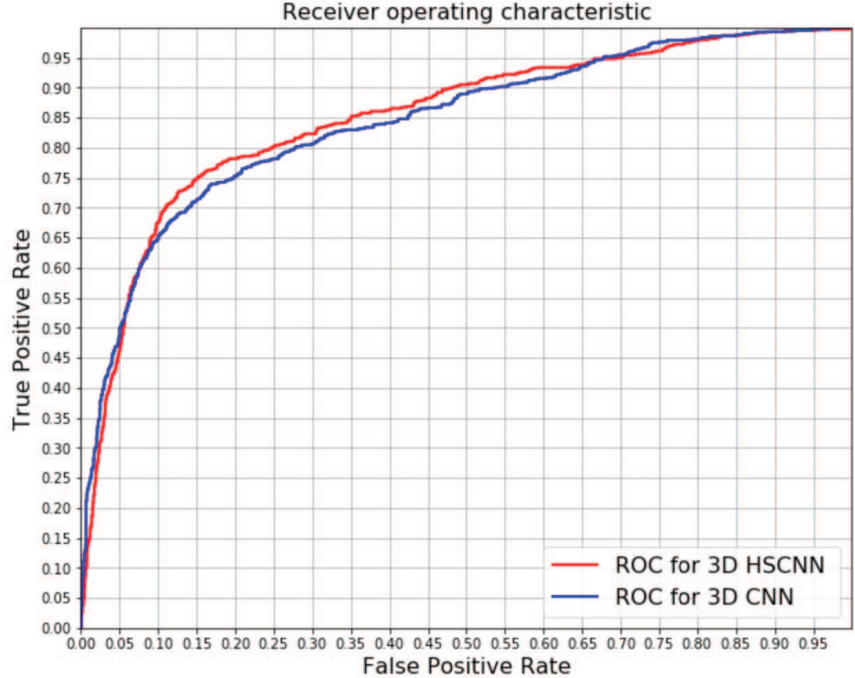} 
	\caption{Receiver operating characteristic curve comparison: HSCNN versus 3D CNN. The AUC of 3D HSCNN is significantly higher than 3D CNN according to a paired T-test as shown in Table \ref{hscnn_t_test}. }
	\label{f_hscnn_roc}
\end{figure}

\begin{table}[t!]
\centering
\caption{Results comparison: HSCNN versus 3D CNN.}
\label{HSCNN_vs_CNN}
\begin{tabular}{c | cccc}
\hline
\textbf{Model} & \textbf{AUC (SD)} & \textbf{Accuracy (SD)} & \textbf{Sensitivity (SD)} & \textbf{Specificity (SD)} \\ \hline
3D CNN  & 0.847 (0.024) & 0.834  (0.022) & 0.668 (0.040)    & 0.889 (0.022)    \\ \hline
HSCNN                        & 0.856 (0.026) & 0.842 (0.025)  & 0.705 (0.045)    & 0.889 (0.022) \\ \hline
                                   
\end{tabular}
\end{table}

To evaluate and compare the HSCNN performance on lung nodule malignancy prediction, a 3D convolutional neural network (3D\_CNN) was first implemented as a baseline model, shown in Figure \ref{f_baseline_hscnn}b. This 3D CNN uses the same feature learning and high-level task components architecture as the HSCNN but has the low-level task component removed. The baseline model was trained and evaluated using the same 4-fold cross validation process and with the same data splitting for each fold (using the same randomization seed). 

Figure \ref{f_hscnn_roc} shows the receiver operating characteristic (ROC) curve plots comparing HSCNN versus 3D CNN performance. These plots represent the intuitive trade-off between sensitivity and specificity. By visual inspection of the ROC curves, HSCNN performs better than the traditional 3D CNN model. The area under the ROC curve (AUC) quantitatively compares the overall performance of a classification model and is frequently used as a metric to access performance in nodule classification \cite{shen2017multi, ciompi2015automatic, hancock2016lung, clark2013cancer, froz2017lung}. Table \ref{f_hscnn_roc} summarizes the mean AUC score, accuracy, sensitivity, and specificity for both models. The HSCNN model achieved a mean AUC 0.856, mean accuracy 0.842, mean sensitivity 0.705 and mean specificity 0.889; while the 3D CNN model achieved a mean AUC 0.847, mean accuracy 0.834, mean sensitivity 0.668 and mean specificity 0.889. Both ROC plots and metric assessments show that the proposed HSCNN achieved better performance for malignancy prediction compared with the conventional 3D CNN approach. 

To assess the statistical significance of model performance improvements, we conducted a paired sample t-test to evaluate the mean differences in AUC scores between the HSCNN and 3D CNN model. Group 1 consists of the AUC score of the HSCNN model for each holdout test fold during the cross validation. Group 2 consists of the corresponding AUC score for the 3D CNN for the same fold. The null hypothesis is that the mean difference of AUC score between these two models equals 0. Table \ref{hscnn_t_test} summarizes the AUC scores for these groups and results of a paired t-test. The test obtained a p-value of 0.005 and confidence interval of $[0.0051, 0.0129]$, thus rejecting the null hypothesis and indicating that the HSCNN model achieved a statistically significantly better AUC relative to the 3D CNN. The mean improvement of the AUC score was 0.009. This finding demonstrates that adding a low-level task component on an existing CNN structure may improve the prediction of malignancy in a lung nodule. 

\begin{table}[t!]
\centering
\caption{Paired T-Test summarizes for AUC scores between HSCNN and 3D CNN model on test set of each fold. CI represents for confidence interval.}
\label{hscnn_t_test}
\begin{tabular}{c | ccc | c}
\hline
\textbf{Test Fold} & \textbf{\begin{tabular}[c]{@{}c@{}}HSCNN \\ AUC\end{tabular}} & \textbf{\begin{tabular}[c]{@{}c@{}}3D CNN\\ AUC\end{tabular}} & \textbf{\begin{tabular}[c]{@{}c@{}}AUC Difference\\ (HSCNN - 3D\_CNN)\end{tabular}} & \textbf{Paired T-Test}   \\ \hline
Fold 1             & 0.878                                                         & 0.869                                                         & 0.009                                                                               & \multirow{4}{*}{\begin{tabular}[c]{@{}c@{}}P-value=0.005, \\ Mean\_difference=0.009, \\ CI = {[}0.0051, 0.0129{]}\end{tabular}} \\
Fold 2             & 0.813                                                         & 0.807                                                         & 0.006                                                                               &                                                                                                      \\
Fold 3             & 0.874                                                         & 0.862                                                         & 0.012                                                                               &                                                                                                      \\
Fold 4             & 0.860                                                         & 0.851                                                         & 0.009                                                                               &   \\ \hline                                                                               
\end{tabular}
\end{table}

We also compared our results with current deep learning models for lung nodule malignancy prediction, as reported in the literature to date. Kumar et al. \cite{kumar2015lung} developed a deep autoencoder-based model with 4,323 nodules of the LIDC dataset, achieving model accuracy of 0.7501. Hua et al. \cite{hua2015computer} presented a CNN model and deep belief network (DBN) model. Both models were trained and validated using 2,545 lung nodule samples from LIDC. The CNN model had specificity of 0.787 and sensitivity 0.737; and the DBN model obtained specificity of 0.822 and sensitivity 0.734. In \cite{shen2015multi}, Shen et al. used a model based on multi-scale 3D CNN. Developed with 1,375 LIDC nodule samples, the average accuracy is reported above 0.84 with different configurations. In \cite{shen2017multi}, Shen et al. extended this multi-scale model using a multi-crop approach and achieved accuracy of 0.839, 0.8636, and 0.8714 with 340, 1030 and 1375 nodules of LIDC, respectively. All of these previously reported methods were evaluated with only training and validation data splits without an independent holdout test dataset as discussed in Section \ref{sec:method_training}. This evaluation design might lead to information leakage due to the use of the validation data for optimal model parameters selection and may overestimate the model performance. In general, our model achieved better or similar performances compared with these reported methods.

\subsection{Semantic Feature Prediction Results and Model Interpretability}
\label{sec:results_results_semantic}

\begin{table}[t!]
\centering
\caption{Classification performance for semantic feature predictions.}
\label{Semantic_label_performance}
\begin{tabular}{c|cccc}
\hline
\textbf{\begin{tabular}[c]{@{}c@{}}Semantic \\ Features\end{tabular}} & \textbf{Accuracy (SD)} & \textbf{AUC (SD)} & \textbf{Specificity (SD)} & \textbf{Sensitivity (SD)} \\ \hline
Calcification              & 0.908 (0.050)          & 0.930 (0.034)     & 0.763 (0.092)             & 0.930 (0.067)             \\
Margin                     & 0.725 (0.049)          & 0.776 (0.033)     & 0.632 (0.109)             & 0.758 (0.091)             \\
Subtlety                   & 0.719 (0.019)          & 0.803 (0.015)     & 0.796 (0.045)             & 0.673 (0.044)             \\
Texture                    & 0.834 (0.086)          & 0.850 (0.042)     & 0.636 (0.199)             & 0.855 (0.108)             \\
Sphericity                & 0.552 (0.027)          & 0.568 (0.015)     & 0.554 (0.076)             & 0.552 (0.095)            \\ \hline
\end{tabular}
\end{table}

Table \ref{Semantic_label_performance} presents the classification performance for each of the low-level tasks (i.e., semantic features). We achieved mean accuracy of 0.908, 0.725, 0.719, 0.834 and 0.552; mean AUC score of 0.930, 0.776, 0.803, 0.850 and 0.568; mean sensitivity of 0.930, 0.758, 0.673, 0.855 and 0.552; and mean specificity of 0.763, 0.632, 0.796, 0.636 and 0.554 for calcification, margin, subtlety, texture, and sphericity, respectively. These results suggest that the HSCNN model is able to learn feature representations that are predictive of semantic features while simultaneously achieving high performance in predicting nodule malignancy. 

\begin{figure}[!t]
	\centering
	\includegraphics[scale=0.95]{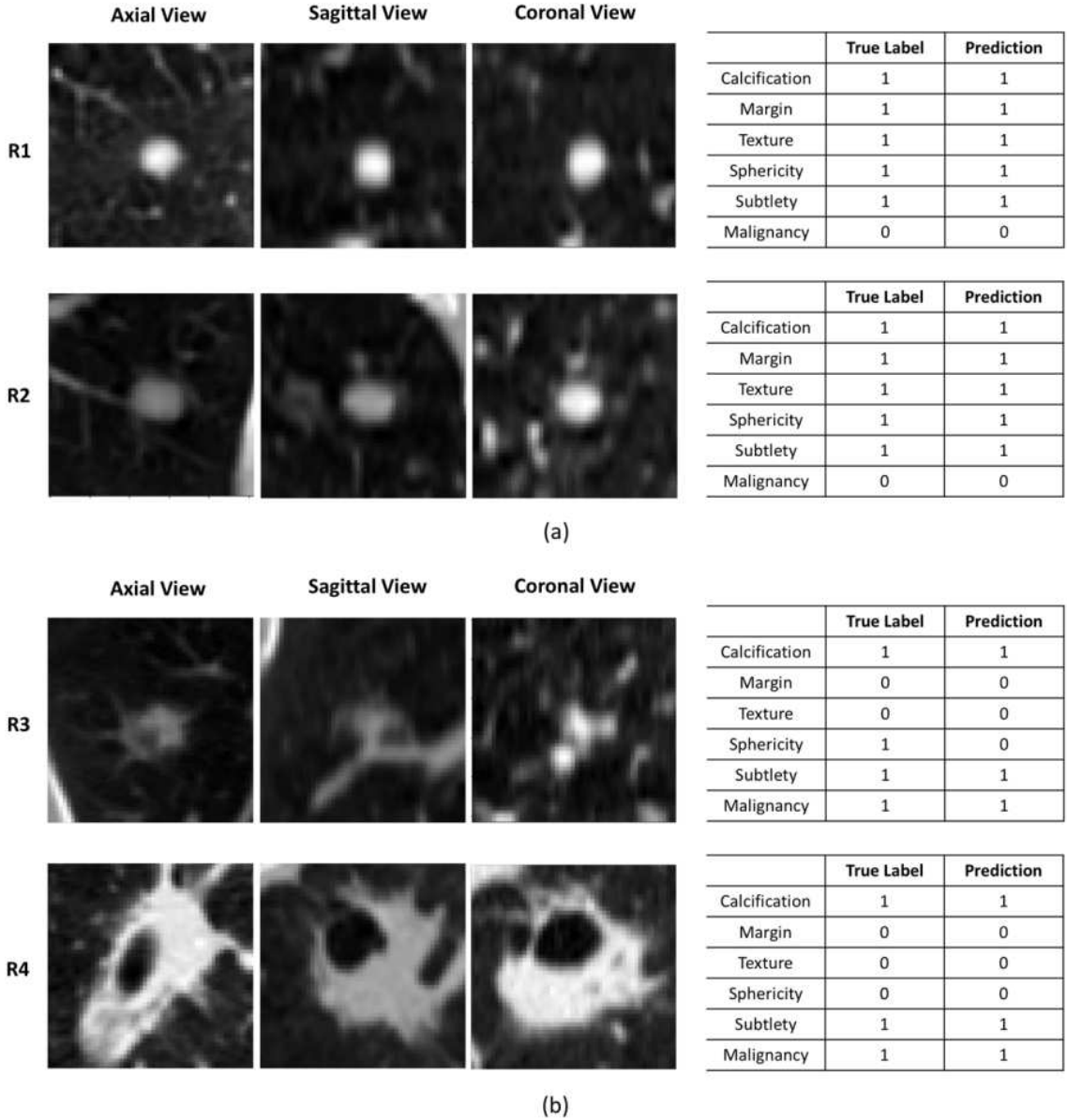} 
	\caption{Illustrating the HSCNN model interpretability: lung nodule central slices, interpretable semantic feature prediction and malignancy prediction. R1, R2, R3 and R4 are four different nodules. (a) Central slices of axial, coronal and sagittal view of two benign nodule samples; true and predicted labels for interpretable semantic features and malignancy. (b) Central slices of axial, coronal and sagittal view of two malignant nodule samples; true and predicted labels for interpretable semantic features and malignancy.}
	\label{f_visual_semantic}
\end{figure}

Figure \ref{f_visual_semantic} demonstrates the interpretability of the HSCNN model by visualizing the central slices of the 3D nodule patches in axial, coronal, and sagittal projections while presenting the predicted interpretable semantic labels along with the malignancy classification results. Figure \ref{f_visual_semantic}a-R1 shows that the HSCNN model classifies the lung nodule as benign (the true label is also benign). This decision correlated to predictions of this nodule as having no calcification, sharp margins, roundness, obvious contrast between nodule and surroundings, and solid consistency. The predictions of these five semantic characteristics are the same as the true label and corresponds to our knowledge about benign lung nodules. Compared to a 3D CNN malignancy prediction model, the HSCNN provides more insight for interpreting its predictions. Similarly, in Figure \ref{f_visual_semantic}b-R3, the proposed model predicts the lung nodule as malignant (true label is also malignant). Different from the benign case, the HSCNN model predicts this nodule having poorly defined margins, ground glass consistency, and non-round shape. This partly explains why the HSCNN makes a malignancy classification with such nodule characteristics corresponding to our expert knowledge about typical malignant nodules. We note that the sphericity predictions made by the model are different from the true label. This result is explained by the fact that while the nodule has a more regular round shape in axial view, the shape is actually more elongated in the two other projections, as shown in Figure \ref{f_visual_semantic}b-R3.

\begin{figure}[!t]
	\centering
	\includegraphics[scale=0.35]{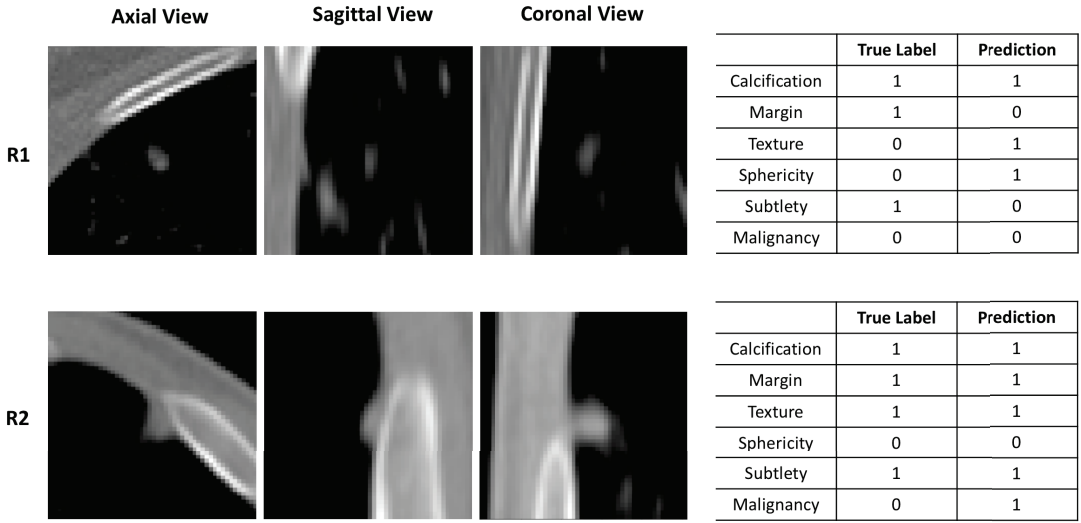} 
	\caption{Example cases where the HSCNN model incorrectly predicts semantic features or cancer malignancy. R1 and R2 are two different nodules. R1: This case has four incorrect semantic feature predictions yet a correct malignancy prediction. R2: This case has all correct semantic predictions yet an incorrect malignancy prediction.}
	\label{f_visual_semantic_incorrect}
\end{figure}

Figure \ref{f_visual_semantic_incorrect} shows two representative cases where the HSCNN fails to predict either one or more semantic features or cancer malignancy. Figure \ref{f_visual_semantic_incorrect}-R1 shows that the HSCNN model classifies the lung nodule correctly as benign but incorrectly for four semantic features of this nodule (margin, texture, sphericity and subtlety). In Figure \ref{f_visual_semantic_incorrect}-R2, the HSCNN model incorrectly classifies the lung nodule as malignant (the true label is benign). However, all semantic features of this nodule are predicted correctly. These two cases present the situation where the correctness are inconsistent between the malignancy and semantic predictions. Section \ref{discuss_results} provides further discussion about how the HSCNN model can be augmented with more semantic features.

\section{Discussion}
\label{discuss_results}

We present a HSCNN model that incorporates domain knowledge into the model architecture design, predicting semantic nodule characteristics along with the primary task of nodule malignancy diagnosis. Five semantic features were considered: calcification, margin, subtlety, texture, and sphericity. Our results in Section \ref{sec:results_results_semantic} suggest that the HSCNN model quantifies these nodule characteristics in a fully data-driven way yet in one integrated model that predicts nodule malignancy. The semantic labels are useful in interpreting the model's predictions for malignancy, mapping the features used by the network to predict the high-level task with established domain knowledge about pulmonary nodules. Section \ref{sec:results_results_malig} shows that this design of the HSCNN also improves the model performance for lung cancer malignancy classification. This study focuses mainly on exploring the values that are added by incorporating the low-level semantic task component design into the CNN architecture. Further optimization of the network architecture to achieve higher prediction performance can be performed. For instance, densely connected designs \cite{huang2016densely} and residual designs \cite{he2016deep} could be used to potentially improve model performance. But due to limitations of computation power, not all designs are optimally searched; we will investigate these as part of future work.

There are some limitations to this study. Our semantic labels did not include those of known higher association with malignancy, such as nodule size, margin spiculation, lobulation, and anatomic location, which have previously been reported as informative \cite{mcwilliams2013probability, swensen1997probability}. In the case of lobulation and spiculation, known labeling errors in the LIDC dataset made them unsuitable for our use. Additionally, semantic labels are subject to moderate inter-reader variability; performance might be enhanced by limiting semantic labels to those on which there is high reader agreement. Third, the malignancy labels provided in the LIDC dataset do not reflect pathological diagnosis but rather, suspicion levels of the interpreting radiologists. Finally, the original semantic features have scales of 5 or 6; binarizing the labels may lose some of the semantic information. Changing the threshold for binary classification would also affect results. Our rationale for binary labels in this case was to overcome data sparsity, where the number of cases labeled for certain scales might be very small compared with the other scales (e.g., only 11 cases are labeled as linear for sphericity out of total 4252 cases). Moreover, analysis shows that inter-reader agreement is much lower for 5 or 6 scales compared with the proposed binary labels. Thus, binary labeling helps to reduce labeling noise caused by inter-reader variability. These limitations may be circumvented by training on large datasets that have been systematically annotated using a shared lexicon that includes discriminating features. In future work we will explore model improvement by including discriminating semantic features, and investigate model variability using different semantic labeling schemes. Although  only five subtask  modules are presented for the  HSCNN architecture in this paper, the HSCNN framework and global loss function could be easily extended to increase or decrease the number of low-level semantic features. This study also paves the way to apply this idea in other disease domain to build interpretable models.

\section{Conclusion}
\label{conclustion_hscnn}

In this paper, we have developed a novel radiologist-interpretable HSCNN model for predicting lung cancer in CT-detected indeterminate nodules. This model is able to simultaneously predict nodule malignancy while classifying five nodule semantic characteristics, including calcification, margin, subtlety, texture, and sphericity of nodules. These diagnostic semantic features predictions are intermediate outputs associated with the final malignancy prediction, and are useful to explain the diagnosis prediction. Information from each low-level semantic feature prediction is incorporated into the malignancy prediction task by employing jump connections. This framework is able to enforce the shared basic convolution modules in the HSCNN to learn features that are generalizable across tasks. This unified model is trained by minimizing a joint global loss function, where the losses of both malignancy and semantic feature prediction tasks are incorporated. Extensive experiments and statistical tests show that the proposed HSCNN model is able to significantly improve the classification performance for nodule malignancy prediction and the semantic characteristics predictions have improved the model interpretability. This trained model could also serve as a lung nodule semantic feature generator.

\section*{Author Contributions}

All authors contributed to the development of the project. SS developed the methodology, conducted the experiments and wrote the manuscript. SXH contributed to the experiments. AATB and DRA provided valuable advice and domain input.  WH provided oversight over the project and contributed to its design. All authors reviewed the manuscript.

\section*{Acknowledgement}

The authors acknowledge the National Cancer Institute and the Foundation for the National Institutes of Health, and their critical role in the creation of the free publicly available LIDC/IDRI Database used in this study. Research reported in this publication was partly supported by the National Cancer Institute of the National Institutes of Health under award number R01 CA210360, the Center for Domain-Specific Computing (CDSC) funded by the NSF Expedition in Computing Award CCF-0926127, and the National Science Foundation under Grant No. 1722516. Computing resources were provided by the NIH Data Commons Pilot and a donation of a TITAN Xp graphics card by the NVIDIA Corporation. The content is solely the responsibility of the authors and does not necessarily represent the official views of sponsor agencies. 

\pagebreak

\section* {\refname}
\bibliographystyle{model1-num-names}
\bibliography{lung_CNN}

\begin{thebibliography}{45}
\expandafter\ifx\csname natexlab\endcsname\relax\def\natexlab#1{#1}\fi
\providecommand{\bibinfo}[2]{#2}
\ifx\xfnm\relax \def\xfnm[#1]{\unskip,\space#1}\fi
\bibitem[{Torre et~al.(2016)Torre, Siegel, and Jemal}]{torre2016lung}
\bibinfo{author}{L.~A. Torre}, \bibinfo{author}{R.~L. Siegel},
  \bibinfo{author}{A.~Jemal},
\newblock \bibinfo{title}{Lung cancer statistics},
\newblock in: \bibinfo{booktitle}{Lung Cancer and Personalized Medicine},
  \bibinfo{publisher}{Springer}, \bibinfo{year}{2016}, pp.
  \bibinfo{pages}{1--19}.
\bibitem[{Shen et~al.(2017)Shen, Han, Petousis, Weiss, Meng, Bui, and
  Hsu}]{shen2017bayesian}
\bibinfo{author}{S.~Shen}, \bibinfo{author}{S.~X. Han},
  \bibinfo{author}{P.~Petousis}, \bibinfo{author}{R.~E. Weiss},
  \bibinfo{author}{F.~Meng}, \bibinfo{author}{A.~A. Bui},
  \bibinfo{author}{W.~Hsu},
\newblock \bibinfo{title}{A bayesian model for estimating multi-state disease
  progression},
\newblock \bibinfo{journal}{Computers in biology and medicine}
  \bibinfo{volume}{81} (\bibinfo{year}{2017}) \bibinfo{pages}{111--120}.
\bibitem[{Team et~al.(2011)}]{national2011reduced}
\bibinfo{author}{N.~L. S. T.~R. Team}, et~al.,
\newblock \bibinfo{title}{Reduced lung-cancer mortality with low-dose computed
  tomographic screening},
\newblock \bibinfo{journal}{N Engl J Med} \bibinfo{volume}{2011}
  (\bibinfo{year}{2011}) \bibinfo{pages}{395--409}.
\bibitem[{ten Haaf et~al.(2017)ten Haaf, Jeon, Tammem{\"a}gi, Han, Kong,
  Plevritis, Feuer, de~Koning, Steyerberg, and Meza}]{ten2017risk}
\bibinfo{author}{K.~ten Haaf}, \bibinfo{author}{J.~Jeon},
  \bibinfo{author}{M.~C. Tammem{\"a}gi}, \bibinfo{author}{S.~S. Han},
  \bibinfo{author}{C.~Y. Kong}, \bibinfo{author}{S.~K. Plevritis},
  \bibinfo{author}{E.~J. Feuer}, \bibinfo{author}{H.~J. de~Koning},
  \bibinfo{author}{E.~W. Steyerberg}, \bibinfo{author}{R.~Meza},
\newblock \bibinfo{title}{Risk prediction models for selection of lung cancer
  screening candidates: A retrospective validation study},
\newblock \bibinfo{journal}{PLoS medicine} \bibinfo{volume}{14}
  (\bibinfo{year}{2017}) \bibinfo{pages}{e1002277}.
\bibitem[{Zhao et~al.(2013)Zhao, Tan, Bell, Marley, Guo, Mann, Scott, Schwartz,
  and Ghiorghiu}]{zhao2013exploring}
\bibinfo{author}{B.~Zhao}, \bibinfo{author}{Y.~Tan}, \bibinfo{author}{D.~J.
  Bell}, \bibinfo{author}{S.~E. Marley}, \bibinfo{author}{P.~Guo},
  \bibinfo{author}{H.~Mann}, \bibinfo{author}{M.~L. Scott},
  \bibinfo{author}{L.~H. Schwartz}, \bibinfo{author}{D.~C. Ghiorghiu},
\newblock \bibinfo{title}{Exploring intra-and inter-reader variability in
  uni-dimensional, bi-dimensional, and volumetric measurements of solid tumors
  on ct scans reconstructed at different slice intervals},
\newblock \bibinfo{journal}{European journal of radiology} \bibinfo{volume}{82}
  (\bibinfo{year}{2013}) \bibinfo{pages}{959--968}.
\bibitem[{Armato et~al.(2003)Armato, Altman, Wilkie, Sone, Li, Roy
  et~al.}]{armato2003automated}
\bibinfo{author}{S.~G. Armato}, \bibinfo{author}{M.~B. Altman},
  \bibinfo{author}{J.~Wilkie}, \bibinfo{author}{S.~Sone},
  \bibinfo{author}{F.~Li}, \bibinfo{author}{A.~S. Roy}, et~al.,
\newblock \bibinfo{title}{Automated lung nodule classification following
  automated nodule detection on ct: A serial approach},
\newblock \bibinfo{journal}{Medical Physics} \bibinfo{volume}{30}
  (\bibinfo{year}{2003}) \bibinfo{pages}{1188--1197}.
\bibitem[{Shen et~al.(2015)Shen, Bui, Cong, and Hsu}]{shen2015automated}
\bibinfo{author}{S.~Shen}, \bibinfo{author}{A.~A. Bui},
  \bibinfo{author}{J.~Cong}, \bibinfo{author}{W.~Hsu},
\newblock \bibinfo{title}{An automated lung segmentation approach using
  bidirectional chain codes to improve nodule detection accuracy},
\newblock \bibinfo{journal}{Computers in biology and medicine}
  \bibinfo{volume}{57} (\bibinfo{year}{2015}) \bibinfo{pages}{139--149}.
\bibitem[{Duggan et~al.(2015)Duggan, Bae, Shen, Hsu, Bui, Jones, Glavin, and
  Vese}]{duggan2015technique}
\bibinfo{author}{N.~Duggan}, \bibinfo{author}{E.~Bae},
  \bibinfo{author}{S.~Shen}, \bibinfo{author}{W.~Hsu},
  \bibinfo{author}{A.~Bui}, \bibinfo{author}{E.~Jones},
  \bibinfo{author}{M.~Glavin}, \bibinfo{author}{L.~Vese},
\newblock \bibinfo{title}{A technique for lung nodule candidate detection in ct
  using global minimization methods},
\newblock in: \bibinfo{booktitle}{International Workshop on Energy Minimization
  Methods in Computer Vision and Pattern Recognition},
  \bibinfo{organization}{Springer}, pp. \bibinfo{pages}{478--491}.
\bibitem[{Firmino et~al.(2016)Firmino, Angelo, Morais, Dantas, and
  Valentim}]{firmino2016computer}
\bibinfo{author}{M.~Firmino}, \bibinfo{author}{G.~Angelo},
  \bibinfo{author}{H.~Morais}, \bibinfo{author}{M.~R. Dantas},
  \bibinfo{author}{R.~Valentim},
\newblock \bibinfo{title}{Computer-aided detection (cade) and diagnosis (cadx)
  system for lung cancer with likelihood of malignancy},
\newblock \bibinfo{journal}{Biomedical engineering online} \bibinfo{volume}{15}
  (\bibinfo{year}{2016}) \bibinfo{pages}{2}.
\bibitem[{Amir and Lehmann(2016)}]{amir2016after}
\bibinfo{author}{G.~J. Amir}, \bibinfo{author}{H.~P. Lehmann},
\newblock \bibinfo{title}{After detection:: The improved accuracy of lung
  cancer assessment using radiologic computer-aided diagnosis},
\newblock \bibinfo{journal}{Academic radiology} \bibinfo{volume}{23}
  (\bibinfo{year}{2016}) \bibinfo{pages}{186--191}.
\bibitem[{Huang et~al.(2017)Huang, Park, Yan, Lee, Chu, Lin, Hussien, Rathmell,
  Thomas, Chen et~al.}]{huang2017added}
\bibinfo{author}{P.~Huang}, \bibinfo{author}{S.~Park},
  \bibinfo{author}{R.~Yan}, \bibinfo{author}{J.~Lee}, \bibinfo{author}{L.~C.
  Chu}, \bibinfo{author}{C.~T. Lin}, \bibinfo{author}{A.~Hussien},
  \bibinfo{author}{J.~Rathmell}, \bibinfo{author}{B.~Thomas},
  \bibinfo{author}{C.~Chen}, et~al.,
\newblock \bibinfo{title}{Added value of computer-aided ct image features for
  early lung cancer diagnosis with small pulmonary nodules: A matched
  case-control study},
\newblock \bibinfo{journal}{Radiology} \bibinfo{volume}{286}
  (\bibinfo{year}{2017}) \bibinfo{pages}{286--295}.
\bibitem[{Zinovev et~al.(2011)Zinovev, Feigenbaum, Furst, and
  Raicu}]{zinovev2011probabilistic}
\bibinfo{author}{D.~Zinovev}, \bibinfo{author}{J.~Feigenbaum},
  \bibinfo{author}{J.~Furst}, \bibinfo{author}{D.~Raicu},
\newblock \bibinfo{title}{Probabilistic lung nodule classification with belief
  decision trees},
\newblock in: \bibinfo{booktitle}{Engineering in medicine and biology society,
  EMBC, 2011 annual international conference of the IEEE},
  \bibinfo{organization}{IEEE}, pp. \bibinfo{pages}{4493--4498}.
\bibitem[{Way et~al.(2009)Way, Sahiner, Chan, Hadjiiski, Cascade, Chughtai,
  Bogot, and Kazerooni}]{way2009computer}
\bibinfo{author}{T.~W. Way}, \bibinfo{author}{B.~Sahiner},
  \bibinfo{author}{H.-P. Chan}, \bibinfo{author}{L.~Hadjiiski},
  \bibinfo{author}{P.~N. Cascade}, \bibinfo{author}{A.~Chughtai},
  \bibinfo{author}{N.~Bogot}, \bibinfo{author}{E.~Kazerooni},
\newblock \bibinfo{title}{Computer-aided diagnosis of pulmonary nodules on ct
  scans: Improvement of classification performance with nodule surface
  features},
\newblock \bibinfo{journal}{Medical physics} \bibinfo{volume}{36}
  (\bibinfo{year}{2009}) \bibinfo{pages}{3086--3098}.
\bibitem[{Shen et~al.(2015)Shen, Zhou, Yang, Yang, and Tian}]{shen2015multi}
\bibinfo{author}{W.~Shen}, \bibinfo{author}{M.~Zhou},
  \bibinfo{author}{F.~Yang}, \bibinfo{author}{C.~Yang},
  \bibinfo{author}{J.~Tian},
\newblock \bibinfo{title}{Multi-scale convolutional neural networks for lung
  nodule classification},
\newblock in: \bibinfo{booktitle}{International Conference on Information
  Processing in Medical Imaging}, \bibinfo{organization}{Springer}, pp.
  \bibinfo{pages}{588--599}.
\bibitem[{Shen et~al.(2017)Shen, Zhou, Yang, Yu, Dong, Yang, Zang, and
  Tian}]{shen2017multi}
\bibinfo{author}{W.~Shen}, \bibinfo{author}{M.~Zhou},
  \bibinfo{author}{F.~Yang}, \bibinfo{author}{D.~Yu},
  \bibinfo{author}{D.~Dong}, \bibinfo{author}{C.~Yang},
  \bibinfo{author}{Y.~Zang}, \bibinfo{author}{J.~Tian},
\newblock \bibinfo{title}{Multi-crop convolutional neural networks for lung
  nodule malignancy suspiciousness classification},
\newblock \bibinfo{journal}{Pattern Recognition} \bibinfo{volume}{61}
  (\bibinfo{year}{2017}) \bibinfo{pages}{663--673}.
\bibitem[{Piedra et~al.(2016)Piedra, Taira, El-Saden, Ellingson, Bui, and
  Hsu}]{piedra2016assessing}
\bibinfo{author}{E.~A.~R. Piedra}, \bibinfo{author}{R.~K. Taira},
  \bibinfo{author}{S.~El-Saden}, \bibinfo{author}{B.~M. Ellingson},
  \bibinfo{author}{A.~A. Bui}, \bibinfo{author}{W.~Hsu},
\newblock \bibinfo{title}{Assessing variability in brain tumor segmentation to
  improve volumetric accuracy and characterization of change},
\newblock in: \bibinfo{booktitle}{Biomedical and Health Informatics (BHI), 2016
  IEEE-EMBS International Conference on}, \bibinfo{organization}{IEEE}, pp.
  \bibinfo{pages}{380--383}.
\bibitem[{Ciompi et~al.(2015)Ciompi, de~Hoop, van Riel, Chung, Scholten,
  Oudkerk, de~Jong, Prokop, and van Ginneken}]{ciompi2015automatic}
\bibinfo{author}{F.~Ciompi}, \bibinfo{author}{B.~de~Hoop},
  \bibinfo{author}{S.~J. van Riel}, \bibinfo{author}{K.~Chung},
  \bibinfo{author}{E.~T. Scholten}, \bibinfo{author}{M.~Oudkerk},
  \bibinfo{author}{P.~A. de~Jong}, \bibinfo{author}{M.~Prokop},
  \bibinfo{author}{B.~van Ginneken},
\newblock \bibinfo{title}{Automatic classification of pulmonary peri-fissural
  nodules in computed tomography using an ensemble of 2d views and a
  convolutional neural network out-of-the-box},
\newblock \bibinfo{journal}{Medical image analysis} \bibinfo{volume}{26}
  (\bibinfo{year}{2015}) \bibinfo{pages}{195--202}.
\bibitem[{Kumar et~al.(2015)Kumar, Wong, and Clausi}]{kumar2015lung}
\bibinfo{author}{D.~Kumar}, \bibinfo{author}{A.~Wong}, \bibinfo{author}{D.~A.
  Clausi},
\newblock \bibinfo{title}{Lung nodule classification using deep features in ct
  images},
\newblock in: \bibinfo{booktitle}{Computer and Robot Vision (CRV), 2015 12th
  Conference on}, \bibinfo{organization}{IEEE}, pp. \bibinfo{pages}{133--138}.
\bibitem[{Hua et~al.(2015)Hua, Hsu, Hidayati, Cheng, and
  Chen}]{hua2015computer}
\bibinfo{author}{K.-L. Hua}, \bibinfo{author}{C.-H. Hsu},
  \bibinfo{author}{S.~C. Hidayati}, \bibinfo{author}{W.-H. Cheng},
  \bibinfo{author}{Y.-J. Chen},
\newblock \bibinfo{title}{Computer-aided classification of lung nodules on
  computed tomography images via deep learning technique},
\newblock \bibinfo{journal}{OncoTargets and therapy} \bibinfo{volume}{8}
  (\bibinfo{year}{2015}).
\bibitem[{Farag et~al.(2011)Farag, Ali, Graham, Farag, Elshazly, and
  Falk}]{farag2011evaluation}
\bibinfo{author}{A.~Farag}, \bibinfo{author}{A.~Ali},
  \bibinfo{author}{J.~Graham}, \bibinfo{author}{A.~Farag},
  \bibinfo{author}{S.~Elshazly}, \bibinfo{author}{R.~Falk},
\newblock \bibinfo{title}{Evaluation of geometric feature descriptors for
  detection and classification of lung nodules in low dose ct scans of the
  chest},
\newblock in: \bibinfo{booktitle}{Biomedical Imaging: From Nano to Macro, 2011
  IEEE International Symposium on}, \bibinfo{organization}{IEEE}, pp.
  \bibinfo{pages}{169--172}.
\bibitem[{Lin et~al.(2013)Lin, Huang, Lee, and Wu}]{lin2013automatic}
\bibinfo{author}{P.-L. Lin}, \bibinfo{author}{P.-W. Huang},
  \bibinfo{author}{C.-H. Lee}, \bibinfo{author}{M.-T. Wu},
\newblock \bibinfo{title}{Automatic classification for solitary pulmonary
  nodule in ct image by fractal analysis based on fractional brownian motion
  model},
\newblock \bibinfo{journal}{Pattern Recognition} \bibinfo{volume}{46}
  (\bibinfo{year}{2013}) \bibinfo{pages}{3279--3287}.
\bibitem[{Jorritsma et~al.(2015)Jorritsma, Cnossen, and van
  Ooijen}]{jorritsma2015improving}
\bibinfo{author}{W.~Jorritsma}, \bibinfo{author}{F.~Cnossen},
  \bibinfo{author}{P.~van Ooijen},
\newblock \bibinfo{title}{Improving the radiologist--cad interaction: designing
  for appropriate trust},
\newblock \bibinfo{journal}{Clinical radiology} \bibinfo{volume}{70}
  (\bibinfo{year}{2015}) \bibinfo{pages}{115--122}.
\bibitem[{Kim et~al.(2015)Kim, Park, Goo, Wildberger, and
  Kauczor}]{kim2015quantitative}
\bibinfo{author}{H.~Kim}, \bibinfo{author}{C.~M. Park}, \bibinfo{author}{J.~M.
  Goo}, \bibinfo{author}{J.~E. Wildberger}, \bibinfo{author}{H.-U. Kauczor},
\newblock \bibinfo{title}{Quantitative computed tomography imaging biomarkers
  in the diagnosis and management of lung cancer},
\newblock \bibinfo{journal}{Investigative radiology} \bibinfo{volume}{50}
  (\bibinfo{year}{2015}) \bibinfo{pages}{571--583}.
\bibitem[{Erasmus et~al.(2000)Erasmus, Connolly, McAdams, and
  Roggli}]{erasmus2000solitary}
\bibinfo{author}{J.~J. Erasmus}, \bibinfo{author}{J.~E. Connolly},
  \bibinfo{author}{H.~P. McAdams}, \bibinfo{author}{V.~L. Roggli},
\newblock \bibinfo{title}{Solitary pulmonary nodules: Part i. morphologic
  evaluation for differentiation of benign and malignant lesions},
\newblock \bibinfo{journal}{Radiographics} \bibinfo{volume}{20}
  (\bibinfo{year}{2000}) \bibinfo{pages}{43--58}.
\bibitem[{Kaya and Can(2015)}]{kaya2015weighted}
\bibinfo{author}{A.~Kaya}, \bibinfo{author}{A.~B. Can},
\newblock \bibinfo{title}{A weighted rule based method for predicting
  malignancy of pulmonary nodules by nodule characteristics},
\newblock \bibinfo{journal}{Journal of biomedical informatics}
  \bibinfo{volume}{56} (\bibinfo{year}{2015}) \bibinfo{pages}{69--79}.
\bibitem[{Hancock and Magnan(2016)}]{hancock2016lung}
\bibinfo{author}{M.~C. Hancock}, \bibinfo{author}{J.~F. Magnan},
\newblock \bibinfo{title}{Lung nodule malignancy classification using only
  radiologist-quantified image features as inputs to statistical learning
  algorithms: probing the lung image database consortium dataset with two
  statistical learning methods},
\newblock \bibinfo{journal}{Journal of Medical Imaging} \bibinfo{volume}{3}
  (\bibinfo{year}{2016}) \bibinfo{pages}{044504}.
\bibitem[{Armato et~al.(2011)Armato, McLennan, Bidaut, McNitt-Gray, Meyer,
  Reeves, Zhao, Aberle, Henschke, Hoffmoan et~al.}]{armato2011lung}
\bibinfo{author}{S.~G. Armato}, \bibinfo{author}{G.~McLennan},
  \bibinfo{author}{L.~Bidaut}, \bibinfo{author}{M.~F. McNitt-Gray},
  \bibinfo{author}{C.~R. Meyer}, \bibinfo{author}{A.~P. Reeves},
  \bibinfo{author}{B.~Zhao}, \bibinfo{author}{D.~R. Aberle},
  \bibinfo{author}{C.~I. Henschke}, \bibinfo{author}{E.~A. Hoffmoan}, et~al.,
\newblock \bibinfo{title}{The lung image database consortium (lidc) and image
  database resource initiative (idri): a completed reference database of lung
  nodules on ct scans},
\newblock \bibinfo{journal}{Medical physics} \bibinfo{volume}{38}
  (\bibinfo{year}{2011}) \bibinfo{pages}{915--931}.
\bibitem[{McNitt-Gray et~al.(2007)McNitt-Gray, Armato, Meyer, Reeves, McLennan,
  Pais, Freymann, Brown, Engelmann, Bland et~al.}]{mcnitt2007lung}
\bibinfo{author}{M.~F. McNitt-Gray}, \bibinfo{author}{S.~G. Armato},
  \bibinfo{author}{C.~R. Meyer}, \bibinfo{author}{A.~P. Reeves},
  \bibinfo{author}{G.~McLennan}, \bibinfo{author}{R.~C. Pais},
  \bibinfo{author}{J.~Freymann}, \bibinfo{author}{M.~S. Brown},
  \bibinfo{author}{R.~M. Engelmann}, \bibinfo{author}{P.~H. Bland}, et~al.,
\newblock \bibinfo{title}{The lung image database consortium (lidc) data
  collection process for nodule detection and annotation},
\newblock \bibinfo{journal}{Academic radiology} \bibinfo{volume}{14}
  (\bibinfo{year}{2007}) \bibinfo{pages}{1464--1474}.
\bibitem[{\text{A. P. Reeves, A. M. Biancardi}(2011)}]{lidc_size_2011}
\bibinfo{author}{\text{A. P. Reeves, A. M. Biancardi}}, \bibinfo{title}{The
  lung image database consortium (lidc) nodule size report},
  \bibinfo{howpublished}{http://www.via.cornell.edu/lidc/},
  \bibinfo{year}{2011}. \bibinfo{note}{Release: 2011-10-27-2}.
\bibitem[{Clark et~al.(2013)Clark, Vendt, Smith, Freymann, Kirby, Koppel,
  Moore, Phillips, Maffitt, Pringle et~al.}]{clark2013cancer}
\bibinfo{author}{K.~Clark}, \bibinfo{author}{B.~Vendt},
  \bibinfo{author}{K.~Smith}, \bibinfo{author}{J.~Freymann},
  \bibinfo{author}{J.~Kirby}, \bibinfo{author}{P.~Koppel},
  \bibinfo{author}{S.~Moore}, \bibinfo{author}{S.~Phillips},
  \bibinfo{author}{D.~Maffitt}, \bibinfo{author}{M.~Pringle}, et~al.,
\newblock \bibinfo{title}{The cancer imaging archive (tcia): maintaining and
  operating a public information repository},
\newblock \bibinfo{journal}{Journal of digital imaging} \bibinfo{volume}{26}
  (\bibinfo{year}{2013}) \bibinfo{pages}{1045--1057}.
\bibitem[{Froz et~al.(2017)Froz, de~Carvalho~Filho, Silva, de~Paiva, Nunes, and
  Gattass}]{froz2017lung}
\bibinfo{author}{B.~R. Froz}, \bibinfo{author}{A.~O. de~Carvalho~Filho},
  \bibinfo{author}{A.~C. Silva}, \bibinfo{author}{A.~C. de~Paiva},
  \bibinfo{author}{R.~A. Nunes}, \bibinfo{author}{M.~Gattass},
\newblock \bibinfo{title}{Lung nodule classification using artificial crawlers,
  directional texture and support vector machine},
\newblock \bibinfo{journal}{Expert Systems with Applications}
  \bibinfo{volume}{69} (\bibinfo{year}{2017}) \bibinfo{pages}{176--188}.
\bibitem[{\text{The Cancer Imaging Archive}(2017)}]{lidc_annotation_error}
\bibinfo{author}{\text{The Cancer Imaging Archive}}, \bibinfo{title}{Lung image
  database consortium - reader annotation and markup - annotation and markup
  issues/comments},
  \bibinfo{howpublished}{https://wiki.cancerimagingarchive.net/display/Public/LIDC-IDRI},
  \bibinfo{year}{2017}.
\bibitem[{Ioffe and Szegedy(2015)}]{ioffe2015batch}
\bibinfo{author}{S.~Ioffe}, \bibinfo{author}{C.~Szegedy},
\newblock \bibinfo{title}{Batch normalization: Accelerating deep network
  training by reducing internal covariate shift},
\newblock \bibinfo{journal}{arXiv preprint arXiv:1502.03167}
  (\bibinfo{year}{2015}).
\bibitem[{Krizhevsky et~al.(2012)Krizhevsky, Sutskever, and
  Hinton}]{krizhevsky2012imagenet}
\bibinfo{author}{A.~Krizhevsky}, \bibinfo{author}{I.~Sutskever},
  \bibinfo{author}{G.~E. Hinton},
\newblock \bibinfo{title}{Imagenet classification with deep convolutional
  neural networks},
\newblock in: \bibinfo{booktitle}{Advances in neural information processing
  systems}, pp. \bibinfo{pages}{1097--1105}.
\bibitem[{Srivastava et~al.(2014)Srivastava, Hinton, Krizhevsky, Sutskever, and
  Salakhutdinov}]{srivastava2014dropout}
\bibinfo{author}{N.~Srivastava}, \bibinfo{author}{G.~Hinton},
  \bibinfo{author}{A.~Krizhevsky}, \bibinfo{author}{I.~Sutskever},
  \bibinfo{author}{R.~Salakhutdinov},
\newblock \bibinfo{title}{Dropout: A simple way to prevent neural networks from
  overfitting},
\newblock \bibinfo{journal}{The Journal of Machine Learning Research}
  \bibinfo{volume}{15} (\bibinfo{year}{2014}) \bibinfo{pages}{1929--1958}.
\bibitem[{Glorot and Bengio(2010)}]{glorot2010understanding}
\bibinfo{author}{X.~Glorot}, \bibinfo{author}{Y.~Bengio},
\newblock \bibinfo{title}{Understanding the difficulty of training deep
  feedforward neural networks},
\newblock in: \bibinfo{booktitle}{Proceedings of the thirteenth international
  conference on artificial intelligence and statistics}, pp.
  \bibinfo{pages}{249--256}.
\bibitem[{Kingma and Ba(2014)}]{kingma2014adam}
\bibinfo{author}{D.~P. Kingma}, \bibinfo{author}{J.~Ba},
\newblock \bibinfo{title}{Adam: A method for stochastic optimization},
\newblock \bibinfo{journal}{arXiv preprint arXiv:1412.6980}
  (\bibinfo{year}{2014}).
\bibitem[{Simonyan and Zisserman(2014)}]{Simonyan14c}
\bibinfo{author}{K.~Simonyan}, \bibinfo{author}{A.~Zisserman},
\newblock \bibinfo{title}{Very deep convolutional networks for large-scale
  image recognition},
\newblock \bibinfo{journal}{CoRR} \bibinfo{volume}{abs/1409.1556}
  (\bibinfo{year}{2014}).
\bibitem[{Bergstra and Bengio(2012)}]{bergstra2012random}
\bibinfo{author}{J.~Bergstra}, \bibinfo{author}{Y.~Bengio},
\newblock \bibinfo{title}{Random search for hyper-parameter optimization},
\newblock \bibinfo{journal}{Journal of Machine Learning Research}
  \bibinfo{volume}{13} (\bibinfo{year}{2012}) \bibinfo{pages}{281--305}.
\bibitem[{Abadi et~al.(2016)Abadi, Barham, Chen, Chen, Davis, Dean, Devin,
  Ghemawat, Irving, Isard et~al.}]{abadi2016tensorflow}
\bibinfo{author}{M.~Abadi}, \bibinfo{author}{P.~Barham},
  \bibinfo{author}{J.~Chen}, \bibinfo{author}{Z.~Chen},
  \bibinfo{author}{A.~Davis}, \bibinfo{author}{J.~Dean},
  \bibinfo{author}{M.~Devin}, \bibinfo{author}{S.~Ghemawat},
  \bibinfo{author}{G.~Irving}, \bibinfo{author}{M.~Isard}, et~al.,
\newblock \bibinfo{title}{Tensorflow: A system for large-scale machine
  learning.},
\newblock in: \bibinfo{booktitle}{OSDI}, volume~\bibinfo{volume}{16}, pp.
  \bibinfo{pages}{265--283}.
\bibitem[{Chollet et~al.(2015)}]{chollet2015keras}
\bibinfo{author}{F.~Chollet}, et~al., \bibinfo{title}{Keras},
  \bibinfo{year}{2015}.
\bibitem[{Huang et~al.(2016)Huang, Liu, Weinberger, and van~der
  Maaten}]{huang2016densely}
\bibinfo{author}{G.~Huang}, \bibinfo{author}{Z.~Liu}, \bibinfo{author}{K.~Q.
  Weinberger}, \bibinfo{author}{L.~van~der Maaten},
\newblock \bibinfo{title}{Densely connected convolutional networks},
\newblock \bibinfo{journal}{arXiv preprint arXiv:1608.06993}
  (\bibinfo{year}{2016}).
\bibitem[{He et~al.(2016)He, Zhang, Ren, and Sun}]{he2016deep}
\bibinfo{author}{K.~He}, \bibinfo{author}{X.~Zhang}, \bibinfo{author}{S.~Ren},
  \bibinfo{author}{J.~Sun},
\newblock \bibinfo{title}{Deep residual learning for image recognition},
\newblock in: \bibinfo{booktitle}{Proceedings of the IEEE Conference on
  Computer Vision and Pattern Recognition}, pp. \bibinfo{pages}{770--778}.
\bibitem[{McWilliams et~al.(2013)McWilliams, Tammemagi, Mayo, Roberts, Liu,
  Soghrati, Yasufuku, Martel, Laberge, Gingras
  et~al.}]{mcwilliams2013probability}
\bibinfo{author}{A.~McWilliams}, \bibinfo{author}{M.~C. Tammemagi},
  \bibinfo{author}{J.~R. Mayo}, \bibinfo{author}{H.~Roberts},
  \bibinfo{author}{G.~Liu}, \bibinfo{author}{K.~Soghrati},
  \bibinfo{author}{K.~Yasufuku}, \bibinfo{author}{S.~Martel},
  \bibinfo{author}{F.~Laberge}, \bibinfo{author}{M.~Gingras}, et~al.,
\newblock \bibinfo{title}{Probability of cancer in pulmonary nodules detected
  on first screening ct},
\newblock \bibinfo{journal}{New England Journal of Medicine}
  \bibinfo{volume}{369} (\bibinfo{year}{2013}) \bibinfo{pages}{910--919}.
\bibitem[{Swensen et~al.(1997)Swensen, Silverstein, Ilstrup, Schleck, and
  Edell}]{swensen1997probability}
\bibinfo{author}{S.~J. Swensen}, \bibinfo{author}{M.~D. Silverstein},
  \bibinfo{author}{D.~M. Ilstrup}, \bibinfo{author}{C.~D. Schleck},
  \bibinfo{author}{E.~S. Edell},
\newblock \bibinfo{title}{The probability of malignancy in solitary pulmonary
  nodules: application to small radiologically indeterminate nodules},
\newblock \bibinfo{journal}{Archives of internal medicine}
  \bibinfo{volume}{157} (\bibinfo{year}{1997}) \bibinfo{pages}{849--855}.

\end{thebibliography}

\newpage
\section*{Conflict of Interest Statement}
None Declared.

\end{document}